\documentclass[conference]{IEEEtran}
\IEEEoverridecommandlockouts
\usepackage{cite}
\usepackage{amsmath,amssymb,amsfonts}
\usepackage{graphicx}
\usepackage{textcomp}
\usepackage{xcolor}
\usepackage{times}
\usepackage{array}
\usepackage{multirow}
\usepackage{mathrsfs}
\usepackage{soul}
\usepackage{url}
\usepackage[hidelinks]{hyperref}
\usepackage[utf8]{inputenc}
\usepackage{graphicx}
\usepackage{amsmath}
\usepackage{amsthm}
\usepackage{booktabs}
\usepackage{algorithm}
\usepackage{algorithmic}
\usepackage{amssymb}
\usepackage{bbm}
\usepackage{dsfont}
\newcommand{\ud}[1]{\underline{#1}}
\newcommand{\udb}[1]{\underline{\bf{#1}}}

\def\BibTeX{{\rm B\kern-.05em{\sc i\kern-.025em b}\kern-.08em
    T\kern-.1667em\lower.7ex\hbox{E}\kern-.125emX}}
\begin{document}

\title{Toward Intention Discovery for Early Malice Detection in Bitcoin}

\author{\IEEEauthorblockN{Ling Cheng\IEEEauthorrefmark{1},
Feida Zhu\IEEEauthorrefmark{2}, Yong Wang\IEEEauthorrefmark{3} and
Huiwen Liu\IEEEauthorrefmark{4}}
\IEEEauthorblockA{School of Computing and Information Systems\\
Singapore Management University\\
Singapore\\
Email: \{{\IEEEauthorrefmark{1}lingcheng.2020,
\IEEEauthorrefmark{2}fdzhu,
\IEEEauthorrefmark{3}yongwang,
\IEEEauthorrefmark{4}hwliu.2018}\}@smu.edu.sg
}
\IEEEauthorrefmark{2}Corresponding author}

\maketitle

\begin{abstract}
Bitcoin has been subject to illicit activities more often than probably any other financial assets,  due to the pseudo-anonymous nature of its transacting entities. An ideal detection model is expected to achieve all the three properties of (I) early detection, (II) good interpretability, and (III) versatility for various illicit activities. However, existing solutions cannot meet all these requirements, as most of them heavily rely on deep learning without satisfying interpretability and are only available for retrospective analysis of a specific illicit type.

First we present asset transfer paths, which aim to describe addresses' early characteristics. Next, with a decision tree based strategy for feature selection and segmentation, we split the entire observation period into different segments and encode each as a segment vector. After clustering all these segment vectors, we get the global status vectors, essentially the basic unit to describe the whole intention. Finally, a hierarchical self-attention predictor predicts the label for the given address in real-time. A survival module tells the predictor when to stop and proposes the status sequence, namely intention.

With the type-dependent selection strategy and global status vectors, our model can be applied to detect various illicit activities with strong interpretability. The well-designed predictor and particular loss functions strengthen model's prediction speed and interpretability one step further. Extensive experiments on three real-world datasets show that our proposed algorithm outperforms state-of-the-art methods.  Besides, additional case studies justify our model can not only explain existing illicit patterns but can also find new suspicious characters.
\end{abstract}


\section{Introduction}
\label{sec:intro}
Crypto-currency has emerged as a new class of financial asset with ever-increasing market capital and importance. Together with the growing popularity comes a wide range of cyber-crimes~\cite{A4_,A8_,A16_} including hacking, extortion~\cite{A2_,A6_} and money laundering~\cite{A3_,A7_,A15_,A17_}.  Customary in this domain, these criminal behaviors are referred to as \emph{malicious} behaviors and the addresses committing these behaviors \emph{malicious addresses}, due to the fact that each transacting entity in crypto-currencies is represented as an anonymous address instead of an account bind with a verified identity. 

The detection and diagnosis of malicious address in crypto-currency transactions present greater challenges than fraud detection in traditional financial world for the following three distinguishing characteristics of crypto-currency. 
\begin{figure}
	\centering
	\vspace{-0ex}
	\includegraphics[width=.9\columnwidth, angle=0]{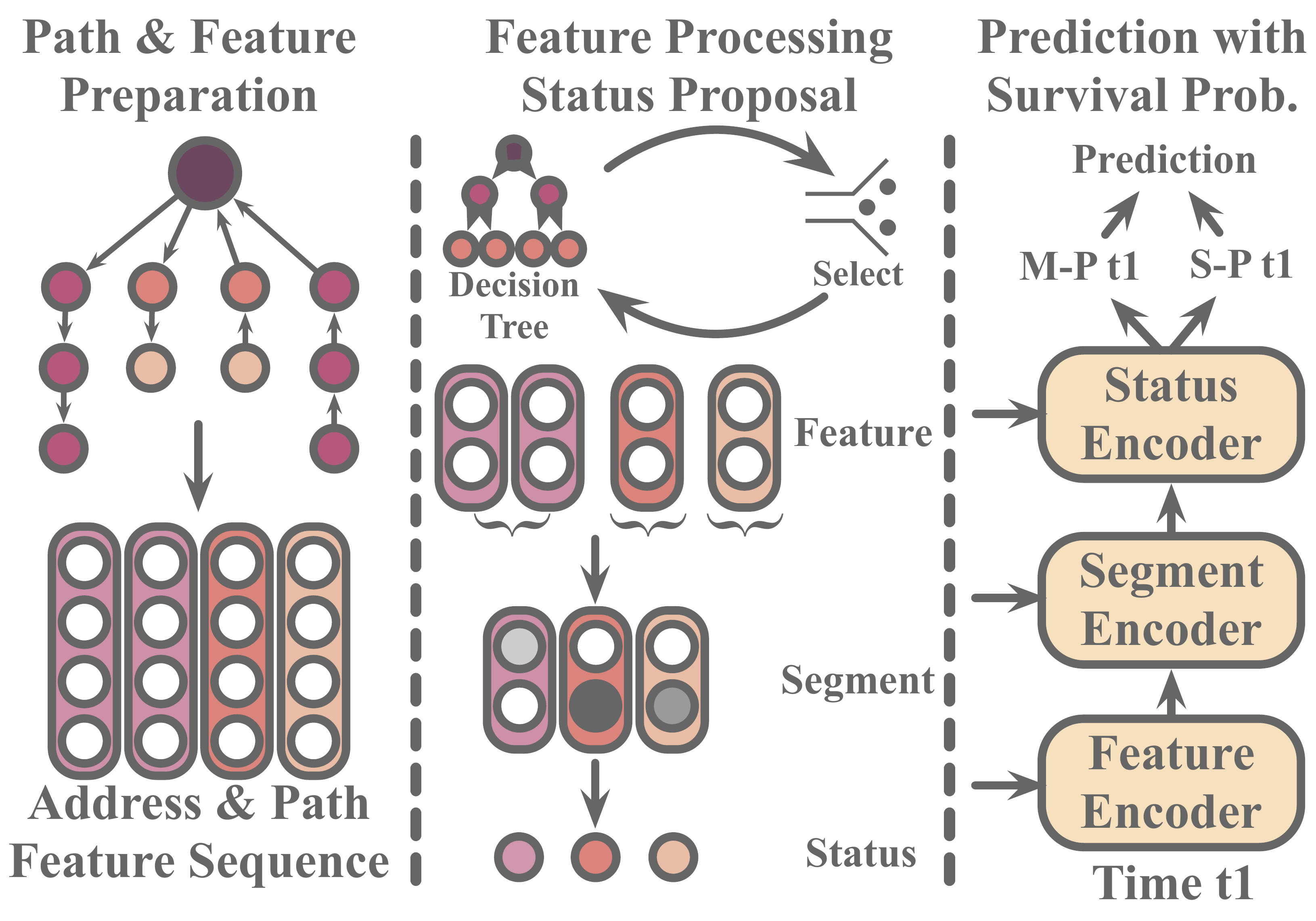}
	\vspace{-0ex}
	\caption{An overview of Intention Monitor. After extracting address and path features, a decision tree based feature selection scheme filters and augments the most significant features. Then, the model splits the feature sequence into several coherent segments and maps them to a set of global statuses through clustering. Finally, a hierarchical transformer with a survival module predicts the address' malicious probability $M-P$ and survival probability $S-P$ in real-time.}
	\label{fig:intro_pipeline}
	\vspace{-4ex}
\end{figure}

\begin{itemize}
\item
\textbf{Early detection is all that matters}. Unlike all other traditional financial assets, crypto-currencies are traded on a  24/7 never-sleeping pace.  
Most malicious behaviors last for a short duration measured only by hours, and will have already inflicted the damage before the associated malicious addresses are forever abandoned if they are not detected in the early stage. Moreover, due to the decentralized nature of crypto-currency's peer-to-peer transactions, retrospective analysis and identification provides little value as 
financial losses are almost impossible to be held back and recovered once the perpetration is completed. 

This challenges most existing graph-based methods as transaction graphs~\cite{A5_,A12_,A13_} needed by these methods must be sufficiently large to provide useful structural information~\cite{A1_}. 
In most cases, the time it takes to form such a transaction graph is much too long to respond effectively to malicious behaviors in action. 
Besides, these methods are usually computationally expensive and time-consuming for early-stage detection.

\item
\textbf{Type-specific features are not versatile enough for malicious behavior detection}. The types of malicious behaviors in crypto-currencies are increasingly diverse, complex, and constantly evolving, ranging from bitcoin-based scams to darknet markets and modus operandi hacking attacks. The characteristics of malicious behaviors also vary a lot across different types.
Manually-engineered features from specific malicious behaviors cannot generalize to other types and unknown ones, let alone apply to other cryptocurrencies in general~\cite{A9_}.
A more general class of features that capture more fundamental characteristics of malicious behaviors across different types is required to achieve the desired versatility. 


\item
\textbf{Interpretability is essential}. Many malicious behaviors in crypto world are packaged as commercial projects to lure victims into investing. Investors must be able to investigate and tell real creditable projects from fraudulent ones. However, most detection methods nowadays rely entirely on deep learning that hardly offers insights into the model's predictions~\cite{A10_}. In particular, most models tend to improve recall at the expense of precision for safety, increasing the risk of missing investment opportunities. From the perspectives of both investors and regulators, model interpretability that offers a deeper understanding of the underlying \emph{intention} behind malicious behaviors is crucial for correct assessment and identification of malicious behaviors. 
\end{itemize}


To address the above-mentioned challenges, we propose Intention Monitor, an early malice detection system based on the notion of \emph{asset transition paths}. The essential idea is based on the fact that, no matter which malicious behavior, the ultimate motive and damage are reflected in significant crypto asset transition between innocent addresses and malicious ones. Patterns extracted from significant asset transition would therefore reveal the intention of malicious behavior across different types. On a high level, our solution progresses in the following 4 stages:

\noindent\textbf{(I) Feature formation.} As shown in Fig.\ref{fig:intro_pipeline}, firstly long-term (LT) and short-term (ST) transition paths, the features of the greatest descriptive power and versatility for early stage malice, would be generated to capture the transaction patterns for both long-term and short-term transition structures.

\noindent\textbf{(II) Feature filtering and temporal assembly.} Secondly, a decision tree-based feature selection would identify features of the best discriminative power for different malicious behavior types. A segmentation strategy is used to assemble features into temporally coherent segments. 


\noindent\textbf{(III) Semantic mapping.} All temporally coherent feature segments then map to a set of global statuses  through clustering. These statuses constitute the semantic units which, after temporal grouping by survival analysis in the next stage, could represent the intentions of malicious behaviors. 

\noindent\textbf{(IV) Model training with intention motif as prediction witness.} Finally, a hierarchical transformer with a survival module is trained to predict malicious address in real-time. The survival module segments the status sequence into intention motifs which serve as witness to the prediction result.

To summarize, the key contributions of this paper are:
\begin{itemize}
    \item
     We propose two novel definitions of asset transfer paths, which are not only effective in capturing Bitcoin transaction patterns for early malice detection, but also applicable to other cryptocurrencies in general, making possible the versatility of the model across different malicious behavior types. In particular, short-term asset transition paths are critical for early stage malice detection.  
    \item
    We provide good interpretability for our malice detection result with intention motif as prediction witness, which is unachievable by those deep-learning dominated models. This is achieved by a combination of (I) a decision tree-based strategy with interval segmentation for feature selection and assembly, and (II) a hierarchical transformer encoder with an survival module to group statuses into sequence of intention motif. 
    \item
     We conduct extensive evaluation and achieve substantially better performance on three different malicious data sets than the state-of-the-art. Furthermore, we present a deep-dive case study on 2017 Binance hack incident to illustrate both the corroborating transaction patterns and unexpected hidden insights for early-sate malice detection that are otherwise unattainable. 
\end{itemize}

\section{Related Work}
\label{sec:related}
On virtual currency trading platforms, there are many crimes involving a large number of addresses. Therefore, detecting the identity information of the address is of great significance to the event's post-analysis and early prediction. Based on the types of features, we divide the existing malicious address detection methods into three categories: (1) Case-related features; (2) General address features; (3) Network-based features.

\subsection{Case-Related features}
Case-Related features model the addresses and activities in a specific event. These detailed analyses are based on the IP addresses of object nodes, public data from exchanges, and labels from related forums. Concretely, some victims provided the criminals' addresses and the detail of the criminal cases they experienced. Except for those detailed data, the time difference between illegal transactions and the sub-structure in criminal cases are also helpful in malice analysis. 
Reid and Harrigan \cite{R1_} combined these topological structures with external IP address information to investigate an alleged theft of Bitcoins.    
To extract information from social media, a transaction-graph-annotation system \cite{R2_} is presented. It matched users with transactions in darknet organizations' activities.
Similarly, by exploiting public social media profile information, in \cite{R3_}, they linked 125 unique users to 20 hidden services, including Pirate Bay and Silk Road.
Marie and Tyler \cite{R4_} presented an empirical analysis of Bitcoin-based scams. By amalgamating reports in online forums, they identified 192 scams and categorized them.
Instead of direct numerical analysis, in \cite{R5_} and \cite{R6_}, they found three anomalous 'worm' structures associated with spam transactions by visualizing the transaction data.
The Case-Related features are often helpful in specific case studies. However, these methods require labor-intensive data collection processing. Besides, most of the insights are only available in particular cases, which can not be generalized to other issues.

\subsection{General address features}
The case-dependent criminal knowledge should be generalized to criminal patterns for a more general detection system.
Many works resort to machine learning for malicious activities and illegal address detection~\cite{A14_}. The first step for a machine learning model is a feature proposal module. Since the transaction is the only possible action for an address, commonly used address features majorly describe related transactions, revealing behavior preferences for the given address. 
Transaction patterns such as transaction time, the index of senders and receivers, and the amount value of transactions can help reveal addresses identity\cite{R7_}.
\cite{R10_} proposed a method for entity classification in Bitcoin. By performing a temporal dissection on BTC, they investigated whether some patterns are repeating in different batches of Bitcoin transaction data. 
On ETH, \cite{R8_} and \cite{R9_} extract features from user accounts and operation codes of the smart contracts to detect latent Ponzi schemes implemented as smart contracts. Considering the intrinsic characteristics of a Ponzi scheme, the extracted features mainly describe the transaction amount, time, and count in a specific period.

\subsection{Network-based features}
Crypto-currency inherently provides a transaction network between addresses. Besides focusing on address-level information, network-based features aim to characterize abnormal addresses from network interaction behaviors. By building an address or transaction network, graph metrics are proven powerful in detecting malicious activities.
By taking advantage of the power degree laws and local outlier factor methods on two Bitcoin transaction graphs, \cite{R13_} detected the most suspicious 30 users, including a justified theft.
\cite{R14_} analyzed the transaction patterns centered around exchanges. In their study, various motifs are introduced in directed hypergraphs, especially a 2-motif as a potential laundering pattern.
\cite{R15_} proposed two kinds of heterogeneous temporal motifs in the Bitcoin transaction network and applied them to detect mixing service addresses. 
EdgeProp \cite{R16_}, a GCN-based model, was proposed to learn the representations of nodes and edges in large-scale transaction networks.
\cite{R17_} analyzed two kinds of random walk-based embedding methods which can encode some specific network features.

\section{Problem Formulation}
\label{sec:prob_form}
\subsection{Problem Definition}
\label{sec:prob_def}
Addresses and transactions are the two major attributes of each Bitcoin transaction record. 
Wallet addresses result from a series of digital signature operations based on the public key, calculated from the private key with the elliptic encryption algorithm.

A BTC transaction $tx$ has a set of inputs $I$= $\{i_1, i_2, \dots i_{|I|}\}$ and a set of outputs $J$= $\{ j_1, j_2, \dots j_{|J|}\}$.
The essences of input and output are still transactions. 
If the funds in this transaction are unspent, then the output will be marked as UTXO (unspent transaction output). 
$tx$ records token distribution between $I$ and $J$. The incoming tokens will flow into a pool and then to the outgoing transactions according to the prior agreement proportion. 
There is no record of how many tokens flow from an Input $i$ to an Output $j$.
Thus, we have to build a complete transaction bipartite graph in $tx$
and can get $|I|\times|J|$ transaction pairs in total. 
In other words, a BTC transaction has $|I|\times|J|$ transaction pairs inside.
Let $D_{t_m}$=$\{d^i_{t_m}\}_{i=1}^N$=$\{(l^i, T_{in, t_m}^{i}, T_{out, t_m}^{i})\}_{i=1}^N$, where $l^i \in \{0, 1\}$ is the label of Address $i$, and $0$ or $1$ stands for regular and malicious addresses respectively.
$T_{in, t_m}^{i}$=$[tx_{in, 1}^{i}, tx_{in, 2}^{i}, \dots tx_{N_{in,t_m}}^{i}]$ is the set of all transactions with Address $i$ as the input address by the $t_m$-th time step, and $T_{out, t_m}^{i}$ is the set of all transactions with Address $i$ as the output address by the $t_m$-th time step.
For ease of understanding, we denote these two transaction sets as $Receive$ set and $Spend$ set respectively.

\vspace{0.2cm}
\noindent\textbf{Early Malicious Address Detection (EMAD).}
Given a set of addresses $A$, and 
$D_{t_m}$ at $t_m$-th time-step, 
the problem is to find a binary classifier $F$ such that 
\begin{equation}
F(d^i_{t_m})=
\begin{cases}
1& \text{if Address $i$ is illicit}\\
0& \text{Otherwise}
\end{cases}.
\end{equation}
In the early detection task, we require the prediction to be consistent and predict the correct label as early as possible.
We denote the confident time as $t_{c}$, where all predictions of Classifier $F$ after $t_{c}$ are consistent. 
The smallest $t_{c}$ is denoted as $t_{f.c}$. Our purpose is to train the classifier and predict the correct label of an address with the smallest $t_{f.c}$.

\subsection{Solution Overview}
\label{sec:solu_overview}


After formulating the problem, we design the core goals of the early malicious address detection model:
(1) Predict the address label at each time step accurately.
(2) Get the smallest $t_{f.c}$ in the early stage.
Inspired by prior research on illegal activity intention encoding\cite{M1_,M2_}, we develop a novel solution, \emph{Intention Monitor}, to achieve the two goals and facilitate the early detection of malicious addresses.

In particular, as shown in Fig.~\ref{fig:intro_pipeline}, we propose asset transfer paths to describe the transition patterns. These asset transfer paths can essentially capture the transaction characteristics and address intentions by tracing the source and destination of every related transaction.

Next is a Decision Tree-based Feature Selection and Augmentation model (DT-SA).
It is deployed to filter and augment the most significant features for different types of malicious behaviors.
Based on the features selected by DT-SA, 
the Status Proposal Module (SPM) divides the early stage into several segments and weight features in every segment with corresponding importance scores. In this way, SPM enhances the significant features one step further.
Then, SPM clusters all the addresses' segment representations and present a set of global status representations.
Each segment now has a corresponding global status, which can explain the behavioral intention of a given address.

Finally, we build \emph{Hierarchical Survival Transformer}, an efficient early malicious address detection framework.
The framework can 
(1) comprehensively encode the relation among features, segments, and status representations to determine the address's label.
(2) determine whether the model has enough information to make confident predictions and avoid the subsequent input noise.
(3) proposes the status sequence to explain the address's intention according to the endpoint given by survival analysis.

In the subsequent sections, we introduce Asset Transfer Path, DT-SA with SPM, and Intention Monitor in Sections IV, V, and VI, respectively.

\section{Asset Transfer Path}
\label{sec:asset_trans_path}
At the early stage of malicious behaviors, 
the address-based network has not grown to the size for a credible prediction.
Instead, the transaction flow can provide critical information during this period.
We design asset transfer paths that consist of significant transactions for the EMAD task.

As mentioned in Sec.~\ref{sec:prob_form}, there are $|I|\times|J|$ transaction pairs in one BTC transaction.
However, not all transaction pairs are helpful for malicious address detection. 
For example, those transaction pairs to collect change tokens provide limited information.
Instead, those important transactions typically constitute a significant portion of the entire transaction amount.
We call such transactions \emph{significant transactions}, and only consider such important transactions.
%
As illustrated in Fig.~\ref{fig:asset_trans_path}, each node stands for a transaction. 
The transaction nodes inside the dashed gray box means the given address participates in these transactions.

In this box, the left node is a transaction where the address receives tokens from other transactions,
namely the address's \emph{Receive Transaction}.
In this single transaction example, there are three inputs, 
contributing $5\%$, $70\%$, and $25\%$ respectively to the total transaction amount. 
Similarly, the right node is a transaction where the address spent its tokens to the outputs.
This transaction is called as the address's \emph{Spend Transaction}.
This transaction involves multiple outputs (with a distribution of $20\%$, $70\%$, and $10\%$ as in this example
We then propose Influence and Trust Transaction Pairs. Both of them are defined based on transaction amount instead of address balance.

\subsection{Influence Transaction Pair} Given an input set $I$= $\{i_1, i_2, \dots i_{|I|}\}$ 
to an Output $j$ and the transaction pair set is $\{{I\rightarrow j}\}$, i.e., $\{{I\rightarrow j}\}$=$\{(i_1,j),(i_2,j),\ldots,(i_{|I|},j)\}$, we define \emph{Influence Transaction Pair} as follows: 
Given an influence activation threshold $\theta$,
$(i_{k},j)$ is called an \emph{influence transaction pair} for Transaction $j$,
if there exists a $k$ ($1\le k\le |I|$) such that the amount of transaction pair $(i_{k},j)$ contributes to at least a certain proportion of the input amount of Transaction $j$, i.e, $\hat{A}(i_{k},j) \ge \theta \times \hat{A}(\{{I\rightarrow j}\})$, where $\hat{A}(\cdot)$ denotes the amount of a transaction pair or the sum of all transaction pairs.

Given an influence transaction pair $(i_{k},j)$, we can conclude that Output $j$ obtains at least a significant amount (based on the threshold) of the asset in this transaction from Input $i_{k}$. Accordingly, given an \emph{Receive Transaction} $j$ for the given address, to trace back to the source of the asset, we proposed \emph{Backward Path} based on \emph{Influence Transaction Pair}. 
Algo. \ref{alg:bk_path} gives the detail to prepare \emph{Backward Paths} that reveal where $j$ obtains the asset.

\subsection{Trust Transaction Pair}
In addition to tracing back to the asset source, we also need to investigate where the asset flows to, i.e., the destination of the asset transfer. To that end, we define \emph{trust transaction pair} as follows:
Given a set of outputs $J$= $\{j_1, j_2, \dots j_{|J|}\}$,
an Input $i$, and the set of all transaction pairs $\{{i\rightarrow J}\}$, 
%
for an Output $j_k$ ($1\le k\le |J|$),
if the Input $i$ transfers at least a certain proportion of its output amount to it, this transaction pair is called a \emph{trust transaction pair} for Input $i$.
It indicates a certain form of trust from $i$ to $j_k$ in asset transfer. 

Given a trust transaction pair $(i,j_k)$, we can conclude that Input $i$ sends a certain degree of the asset to Output $j_k$. 
To trace $i$'s destinations, we also define \emph{Forward Paths} based on \emph{Trust Transaction Pair}. 
The pipeline to construct \emph{Forward Path} is similar to \emph{Backward Path}. The only difference is the tracing direction.

\begin{figure}
	\centering
	\vspace{-0ex}
	\includegraphics[width=.9\columnwidth, angle=0]{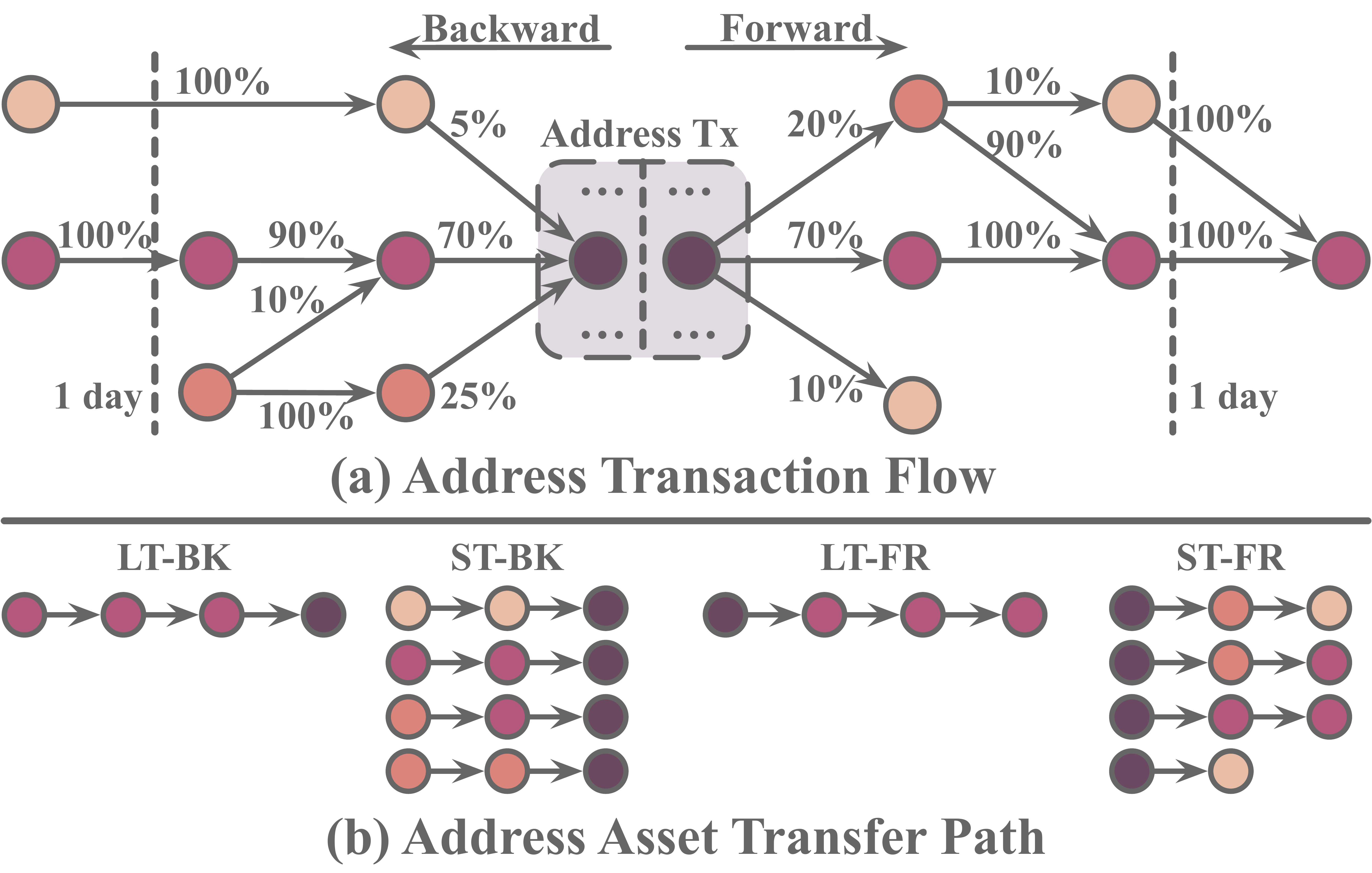}
	\vspace{-0ex}
	\caption{(a) Address transaction flow. Numbers above backward edges are amount proportions to the destination node. 
	Numbers above forward edges are the proportions of the source node. (b) Asset transfer paths. $LT$ and $ST$ stand for long-term and short-term respectively. 
	$FR$ and $BK$ stand for forward and backward respectively.}
	\label{fig:asset_trans_path}
	\vspace{-4ex}
\end{figure}

For brevity, we would refer to both the \emph{Backward Path} and \emph{Forward Path} as \emph{Asset Transfer Paths} and the activation threshold in both directions as \emph{activation threshold}.
To delineate the transaction patterns of an address at both macro and micro levels for both \emph{Backward Path} and \emph{Forward Path}, we define two kinds of time-spans, i.e., long-term and short-term.

 \begin{algorithm}[h]
 \caption{Backward Path Preparation}
 \label{alg:bk_path}
 \begin{algorithmic}[1]
 \renewcommand{\algorithmicrequire}{\textbf{Input:}}
 \renewcommand{\algorithmicensure}{\textbf{Output:}}
 \REQUIRE Initial Output Tx $j^o$, Threshold $\theta$, Time Span $T_{Span}$.
 \ENSURE  Backward Path Set $P$.
 \\ Initialize Backward Path Set $P \gets \{[-,1,j^o]\}$; 
 \\ Initialize Previous hop recorder $P_{pre} \gets \{[-,1,j^o]\}$;
 \\ Initialize Ending Flag $F_{end} \gets False$; 
 \\ $j^o$'s time $T_{j^o} \gets $ time of $j^o$;

  \WHILE{$F_{end} \ne True$}
    \STATE Current hop recorder $P_{now} \gets \{\}$;
    \STATE $F_{end} \gets True$;
    \FOR {$p$ in $P_{pre}$}
    \STATE $j \gets $ Output Tx $p[2]$; 
    \STATE $I \gets $ Input Tx Set of $j$;
    \FOR {$i$ in $I$}
        \STATE $Prop_{i} \gets Amt_{i}/Amt_{I}$;
        \STATE $Score_{i} \gets Prop_{i}*p[1]$; 
        \STATE $T_{i} \gets $ time of $i$;
        \IF {($Score_{i} \ge \theta$ and $T_{j^o} - T_{i}\le T_{Span}$)}
        \STATE Append $[j, Score_{i}, i]$ to $P_{now}$;
        \STATE $F_{end} \gets F_{end}$ $\&\&$ $False$;
        
\ENDIF
\ENDFOR
\ENDFOR
\STATE $P_{pre} \gets P_{now}$;
\STATE $P \gets P \cup P_{pre}$;
\ENDWHILE
\RETURN $P$ 
\end{algorithmic} 
\end{algorithm}

\subsection{Long-Term and Short-Term Path} 
Long-Term (LT) asset transfer paths are those with a larger maximum observation period, e.g., one year as in the current setting. 
We set threshold to 0.5 for long-term paths in this study, since we want to find a pair that takes most of the tokens from its source. 
Take the forward path as an example, if one output transaction's activation score (amount proportion) is larger than 0.5, the remaining transactions' activation scores must be below 0.5.
Then we declare that this transaction pair is the most important one.

The short-term (ST) asset transfer path describes the transition pattern within a short period, e.g., one day. 
Since many unique trading patterns (e.g., pyramid-shaped, pulse-shaped, and spindle-shaped) exist in malicious activities, 
the short-term asset transfer paths reflect the structure of the asset flow.
As the ST path grows, more irrelevant transactions may be included into the path set. 
Thus, the threshold need to increase dynamically to avoid the noise of unrelated transactions.
The threshold for short-term paths is set as $min(\lfloor h/2 \rfloor, 0.9)$, where $h$ is the hop index starting from $0$.
Since the threshold is less than $0.5$ at the beginning, each transaction can have more than one descendant. 
%

%
%

\section{Feature Selection and Augmentation \& Status Proposal Module}
\label{sec:DT_SA_with_SPM}

\begin{algorithm}[ht]
\caption{DT-Based Feature Selection and Augmentation}
\label{alg:DT_SA}
\begin{algorithmic}[1]
\renewcommand{\algorithmicrequire}{\textbf{Input:}}
\renewcommand{\algorithmicensure}{\textbf{Output:}}
\REQUIRE Initial feature list $F^i$, Threshold $\theta$.
\ENSURE Augment, Reserve, and Delete lists $F_A, F_R, F_D$.
\\ Augment and Delete feature list $F_A, F_D \gets \{\}$; 
\\ Reserve feature list $F_R \gets F^i$; 
\\ Average performance score $s_{p}^{A} \gets 0$; 
\\ Best average performance score $s_{p}^{B,A} \gets 0$; 

 \WHILE{$s_{p}^{A} \geq s_{p}^{B,A}$}
   \STATE $s_{p}^{B,A} \gets s_{p}^{A}$
   \STATE $F^{tmp}_{A}, F^{tmp}_{R}, F^{tmp}_{D} \gets F_{A}, F_{R}, F_{D}$;
   \STATE Average performance score $s_{p}^{A} \gets 0$;
   \STATE Best performance score $s_{p}^B \gets 0$;
   \FOR{$idx \gets 1$ to $10$}
       \STATE $DT_{idx}, s_{p, idx} \gets$ Train/Test DT($F_A, F_R, F_D$);
       \STATE $s_{p}^{A} \mathrel{+}= s_{p, idx}/10$;
       \IF{$s_{p, idx}>S^B_p$}
            \STATE Update($F^{tmp}_{A}, F^{tmp}_{R}, F^{tmp}_{D}$)
       \ENDIF
\ENDFOR

\IF {($s_{p}^{A} \geq s_{p}^{B,A}$)}
\STATE $F_A, F_R, F_D \gets F^{tmp}_{A}, F^{tmp}_{R}, F^{tmp}_{D}$;
\ENDIF

\ENDWHILE
\RETURN $F_A, F_R, F_D$ 
\end{algorithmic} 
\end{algorithm}

\begin{figure*}
	\centering
	\vspace{-0ex}
	\includegraphics[width=2.\columnwidth, angle=0]{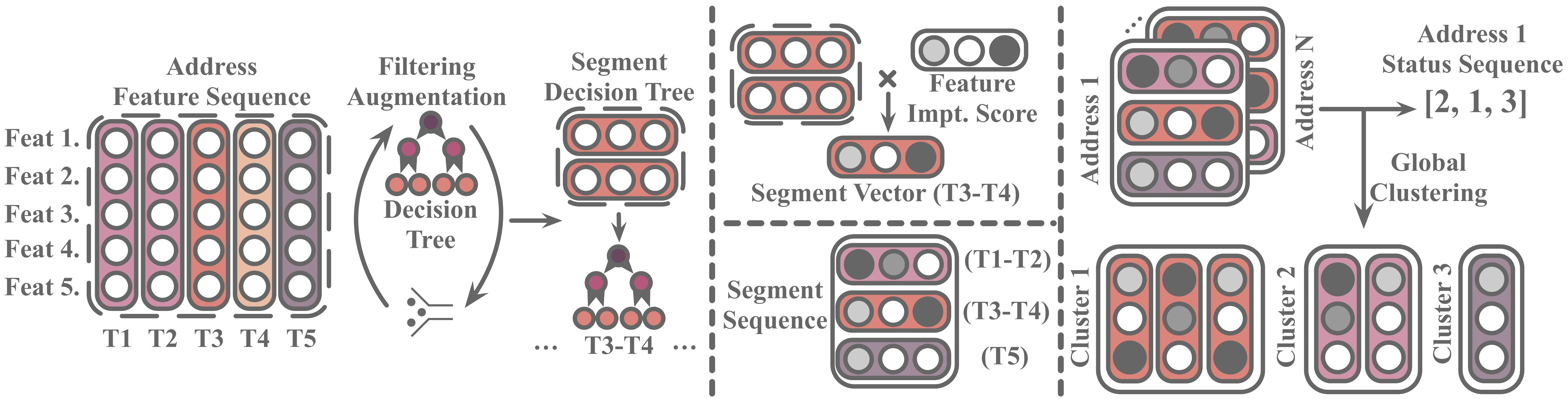}
	\vspace{-0ex}
	\caption{The overview of Decision Tree based feature Selection and Augmentation (DT-SA) and Status Proposal Module (SPM).}
	\label{fig:feat_segm_stat}
	\vspace{-4ex}
\end{figure*}

In this section, we describe the process of Decision Tree-based Feature Selection and Augmentation (DT-SA) and Status Proposal Module (SPM). 
We first introduce the address and asset transfer path features for discriminating malicious addresses. 
Then, we elaborate DT-SA that filters and augments the features for different types of malicious activities.
The state sequence is necessary to understand the address's intention. 
To fetch such a state sequence, we deploy SPM to split the observation period into segments and cluster them to drive global semantic status.
The overview of DT-SA and SPM is shown in Fig.~\ref{fig:feat_segm_stat}.

\subsection{Address \& Transaction Features}
Not all features are useful to describe a specific malicious behavior.
Thus we need to build a larger feature set with stronger representation ability. 
Followed by the \cite{R11_, R12_1, R12_2}, we also use the address features to characterize the address's behaviors.
Considering the runtime complexity, those high computational cost features are skipped.
We extracted 16 address features that characterizes an address from five perspectives:
(1) the current balance of the address, (1 feature)
(2) the number and ratio of spend and receive transactions (so far/recent one hour), (6 features)
(3) the maximum number of spend (receive) transactions per hour, (2 features)
(4) the number of spend (receive) with 0 amount, (2 features)
(5) the time of feature 3 and the time difference between them, (3 features)
(6) the active hour number and activity rate (2 features).

Moreover, the asset path can provide critical information. For a specific path set, we selected 13 path features
(1) path number, (1 feature)
(2) path length (hop/height), (2 features)
(3) maximum (minimum) input (output) amount (quantity) for each node on the path, (8 features)
(4) path's maximum (minimum) activation score.(2 features)
Also, an address has four path sets (LT-BK, ST-BK, LT-FR, ST-FR), a path set has multiple paths and every path has these 13 path features. 
To characterize the overall properties of this path set, 
we calculated the maximum (max), minimum (min), average (avg), and standard deviation (std) values of every feature except the path number. 
Thus, there are 12*4+1=49 path features for a single path set. 
Since we have four path sets (LT-F, LT-B, ST-F, ST-B) (Fig.~\ref{fig:asset_trans_path}), there are 49*4=196 path features in total. 
We can characterize the early behavior of different types of addresses through these address and path features.

We believe the hundreds of extracted address and path features can summarize addresses' early behaviors from several perspectives.
But for a specific type of activity, not all features are equally helpful, and the introduction of irrelevant features can affect the model's performance (will be justified in Sec~\ref{sec:feat_compare}).
Therefore, we need to select the most discriminative features from hundreds of them.

\subsection{DT-based feature Selection and Augmentation}

We develop a decision tree-based feature filter and augmentation module.
In this module we have three feature sets namely augmentation list, reserve list, and deletion list.
This module augments features in the augmentation list,
keeps feature in the reserve list and removes features in the deletion list.

In the initial round, we take each path feature's mean value and address features as seed features. 
So the number of seed features is 13*4+16=68.
We feed these 68 features into the decision tree model and perform $10$ independent training sessions to obtain a group of models.
We sort the feature importance scores of the best-performing model from high to low (the performing score will be elaborated in Sec.~\ref{sec:metrics}). We denote the maximum importance score among all the features as $s_{imp}^M$. 
Given a feature j with a importance score $s_{imp, j}$, 
if $s_{imp, j}$$\geq$$\theta$*$s_{imp}^M$, we append it into the augmentation list and augment it in the next round of training. $\theta$ is the augmentation threshold.
If $0<s_{imp, j}<\theta$*$s_{imp}^M$, we append it into the reserve list, and we will keep it, but will not augment it in the following round of training.
If $s_{imp, j}$=$0$, we append it into the deletion list, and we will delete this feature in the subsequent training process.


In the second round of training, we first augment the features in the augmentation list.
Here, augmentation means not only using the mean value of the feature but also including its maximum, minimum, and standard deviation (only path features are available for augmentation).
Through the augmentation,
we feed the model with more details about the augmented feature, providing more comprehensive information for early malicious address detection.
%
Then, we append reserve features to the input feature list without changing them.
Finally, considering not all the extended features can improve the model's performance in the augmentation process.
Also, the model may pay more attention to the augmented features(Max, Min, Std) than the original feature's average value. 
Thus, if a feature in the current input feature list also appears in the delete list, we then remove it from the input feature list. 
Algo.~\ref{alg:DT_SA} shows the details of DT-SA.

\subsection{Status Proposal Module}
A series of actions can depict the intention of the given address~\cite{M1_}. 
The changes in the feature temporal sequence reflect the address's state evolution and thus can represent its actual intention.
In different states, addresses have different characteristics. Thus, the model should also have different criteria to discriminate addresses' identities.
In addition, to train a decision tree model, the input data should come from a stable state. 
Using different states' data simultaneously will seriously deteriorate the model's performance.

To solve these two problems, we propose the Status Proposal Module (SPM).
First, to characterize the address's different states,
we need to split the entire observation time window into several segments to build a state sequence.
In every segment, features should be stable enough for training a decision tree.
Therefore, we first calculate the features' average change ratio at each time step.
We denote the highest change ratio as $C^H$ and the change ratio at the $j$-th time step as $C_j$.
If $C_j > \theta * C^H$, we add the corresponding time step $j$ to the splitting time list.
After splitting, we train a decision tree for a specific segment due to the different characteristics of different segments. 
Thus, we get a sequence of segments and a group of decision trees, each decision tree is corresponding to a segment.

A series of address states can reflect its intention, and each segment represents a different state of address behavior.
To describe different states of address behavior, we first define \emph{segment representation}, 
to reflect the most crucial features of an address in this segment.
For the $j$-th segment, we record its beginning and end time points as $b_j$ and $e_j$, and the corresponding feature sequence as $[f_{b_j}, f_{b_j+1}, ..., f_{e_j-1}, f_{e_j}]$, where $f_i$ is the feature vector at $i$-th hour.
The feature importance score list of the corresponding decision tree for the $j$-th segment is $S_{imp,j}$
in which each element stands for the importance score for corresponding feature.
Then the segment vector of this period is calculated as:
\begin{equation}
\begin{aligned}
\label{eq:segment}
g^j = S_{imp,j}*\frac{\sum_{i=b_j}^{e_j}f_{i}}{e_j-b_j}.
\end{aligned}
\end{equation}
%
Our segmentation method ensures the stability of each segment, 
and the decision tree's importance score can reflect the most significant properties in this segment.
Thus, $g^j$ can depict the most informative characteristics of a given address in $j$-th segment.

A segment can only delineate the characteristics of a particular address at a certain interval,
whereas, addresses of the same type may have similar purposes at this interval. 
For example, after the Ransomware addresses are activated, 
most of them will wait for the victims to pay the ransom in a certain segment and transfer it out quickly in another segment.
We can call these two segments as ``waiting segment'' and ``transfer segment'' respectively.

To characterize this general intention, 
we first normalize all addresses' segment vectors, 
then cluster them through the DB-Scan algorithm to obtain status clusters with global semantic meanings such as ``waiting'' or ``transfer''. 
We denote each cluster's center as the status vector.
We then build a status decision tree to indicates the status label for the input normalized segment vector.
%
%
Finally, each address has three vector sequences namely features, segments, and statuses. 

\section{Hierarchical Survival Transformer}
\label{sec:intention_monitor}
As mentioned in SPM, every segment has a corresponding decision tree to predict the address's label based on the information in this segment.
However, this decision tree group fails to utilize temporal patterns, which is extremely useful for depicting the address's intention. 
Besides, as there is no early stopping mechanism in the decision tree group, subsequent redundant noise will make the prediction inconsistent.
The Hierarchical Survival Transformer can encode such temporal patterns and prevent noise with a survival prediction module.

\subsection{Hierarchical Transformer Encoder}
For the $p$-th segment, we denote the beginning and ending time points as $b^p$ and $e^p$, respectively. 
In this segment, we have $e^p$-$b^p$+1 features $\{{f}_i\}^{e^p}_{i=b^p}~({f}_{i}\in \mathbb{R}^{d})$, 
one segment vector $g^p \in \mathbb{R}^{d}$ and a one-hot vector of status $o^p$, where $d$ is the feature dimension. We transform the one-hot vector into a learnable embedding vector via embedding:
\begin{equation}
u^p = \mbox{Embed}(o^p).
\end{equation}

First, we expect to select the most representative information from the feature list, 
so we use an attention layer to focus on those significant time steps. 
\begin{equation}
f^p = \sum_{i=b^p}^{e^p}{\alpha}_{i}f_i,
\end{equation}
\begin{equation}
({\alpha}_{b^p}, \cdots, {\alpha}_{i}, \cdots,  {\alpha}_{e^p})= \mbox{Softmax}(a_{b^p}, \cdots, a_{i}, \cdots, a_{e^p}),
\end{equation} 
\begin{equation}
a_{i}= W^{a}\mbox{tanh}(W^{f,u}[f_i, u^p]),
\end{equation}
where $f^p$ is the weighted feature vector, $W^{f,u} \in \mathbb{R}^{{d}\times{2d}}$ and $W^{a} \in \mathbb{R}^{{d}\times{d}}$ are learnable matrices, and they project 
the concatenation of feature and status vector into the attention calculation space. 

As mentioned before, the information in a single segment can hardly provide enough information for an accurate prediction.
Thus, we need to encode the temporal behaviors along the segment sequence.
%
The three kinds of representation vectors have semantic meanings at different levels, so we apply a self-attention encoder to each of them. 
Each representation's self-attention encoder contains $h$ parallel independent heads.
%
Take feature-level self-attention as an example.
With multi-head self-attention, 
the feature list $F^p$ for the $p$-th segment can be defined as follows: 
\begin{equation}
\begin{aligned}
\label{eq:multi-head-attention}
{F^p} &= \{\hat{f}^i\}^{p}_{i=1} = Concat({H^p_1},\cdots,{H^p_h})W^{O}, \\
{H^p_i}&=Softmax(\frac{({Q}{W^{Q}_{i}}){({K}{W^{K}_{i}})}^T}{\sqrt{d}}){{V}{W^{V}_{i}}},
\end{aligned}
\end{equation}
where ${Q}$ is a matrix of $n_q$ query vectors, ${K}$ and ${V}$ contain $n_k$ keys and values with the same number of dimensions,
and 
$d$ is a scaling factor the same as the feature dimension. 
$W^{Q}_{i}, W^{K}_{i}, W^{V}_{i} \in \mathbb{R}^{{d}\times\frac{d}{h}}$ are the independent head projection matrices.
$W^{O} \in \mathbb{R}^{{d}\times{d}}$ denotes the linear transformation.
Our encoder consists of $N$ multi-head self-attention blocks, ${Q}$ = ${K}$ = ${V}$ = $\{{f}^i\}^{p}_{i=1}$, and $n_q$=$n_k$=$p$.

Even when two addresses stay at the same status, their feature and segment may differ with each other.
To highlight these differences, 
we use concrete representations (i.e., feature and segment vectors) to guide self-attention in the status encoder.
For the three independent parallel self-attention encoders, we bridge them with a fully connected layer.
Specifically,
for both the segment self-attention encoder and status self-attention encoder,
${Q}$ = ${K}$ = ${V}$ = $\{\tilde{g}^i\}^{p}_{i=1}$.
$\tilde{g}^i$ and $\tilde{u}^i$ is given by:
\begin{equation}
\begin{aligned}
\label{eq:multi-head-attention}
\tilde{g}^i &= W^g\mbox{tanh}(W^{f,g}[{g}^i, \hat{f}^i]), \\
\tilde{u}^i &= W^u\mbox{tanh}(W^{g,u}[{u}^i, \hat{g}^i]),
\end{aligned}
\end{equation}
where $[\cdot, \cdot]$ stands for concatenation, $\hat{g}^i$ is the segment representation at the $i$-th split after self-attention.
$W^{f,g}, W^{g,u} \in \mathbb{R}^{{d}\times{2d}}$ and $W^{g}, W^{u} \in \mathbb{R}^{{d}\times{d}}$ are learnable matrices.
After encoding all the three levels of representations, 
the final output vector $\hat{u}^p$ is given by the averaged feature vector of the final self-attended status sequence $U^p$. 
Finally, based on $\hat{u}^p$, the model gives the final prediction of the $p$-th splitting period as:
\begin{equation}
y^p = \mbox{Sigmoid}(W^{l}*\hat{u}^p),
\end{equation}
where $W^{l} \in \mathbb{R}^{{d}\times{1}}$ is a learnable fully connected layer.

\subsection{Survival Prediction Analysis}
The survival function $S(t)$ of an event represents the probability that this event has not occurred by time $t$.
If the model has collected enough information to make a confident prediction, it can just copy the previous prediction or stop predicting.
In this way, the model can have a more consistent performance and also get rid of continuous noise.
Thus, we define the event as ``the model has collected enough information to predict the address label''.

%
As in survival analysis~\cite{E5_}, 
the hazard function $\lambda_{t}$ is the event's instantaneous occurrence rate at time $t$ given that the event does not occur before time $t$. 
Since, the observation time is discrete in our case, we use $t$ to denote a timestamp.
The association between $S(t)$ and $\lambda_{t}$ can be calculated as:
\begin{equation}
\begin{aligned}
\label{eq:dsicrete_survival_function}
S(t)&=P(T\geq{t}) = \sum_{k=t}^{\infty}f(x), \\
\lambda_{t}&=f(t)/S(t),\\
S(t)&= exp({-\sum_{k=1}^{t}\lambda_{k}}).
\end{aligned}
\end{equation}

Since the survival function can only monotonically decrease, a \emph{softplus} function is usually deployed to guarantee the hazard rate $\lambda_{t} = softplus(x_t) = ln(1+exp({x_t}))$ is always positive. 
Here, we calculate the hazard rate as follows:
\begin{equation}
\begin{aligned}
\lambda_{t}&= ln(1+\mbox{exp}({W^{hz}\hat{u}^p})),\\
S(t)&= exp({-\sum_{k=1}^{t}\lambda_k}),
\end{aligned}
\end{equation}
where $W^{hz}$ is the linear projection matrices.
Thus, the final prediction $\hat{y}^t$ at $t$-th split is given by:
\begin{equation}
\hat{y}^{t} = S(t)*y^t + (1-S(t))*\hat{y}^{t-1}.
\end{equation}

The introduction of survival prediction analysis can not only accelerate the prediction process but also group statuses into an intention sequence. 
We denote the time step when the survival probability equals $0$ as $t_{die}$. 
At this time step, the model has collected enough information to predict the address's label. 
In other words, the model has figured out the address's intention. 
Thus, we call the status sequence $\{\hat{u}^i\}^{t_{die}}_{i=1}$ as the addresses' intention sequences.

\subsection{Consistent and Early Boost Loss Function}
For an address $i$ at $t$-th segment, the early detection likelihood that this address is malicious and the negative logarithm prediction $loss^P$ are defined below:
\begin{equation}
\begin{aligned}
likelihood & = (1-\hat{y}^t_i)^{l_i}({\hat{y}^t_i})^{1-l_i},\\
loss^P_{i,t} & = ({l_i}-1)*\mbox{log}({\hat{y}^t_i}) - {l_i}*\mbox{log}(1-{\hat{y}^t_i}).
\end{aligned}
\end{equation}

An accurate and reliable model should provide a consistent prediction.
For an ideal model, the current prediction should be consistent with the previous prediction at every time split.
Thus, we introduce the consistency loss $loss^C$ to improve the predictions' conformity:
\begin{equation}
loss^C_{i, t} =
\begin{cases}
0& sign((\hat{y}^t_i-0.5)*(\hat{y}^{t-1}_i-0.5))>=0, \\
1& else,
\end{cases}
\end{equation}
where $0.5$ is the decision boundary of positive (malicious) and negative (regular).

To accelerate the prediction speed, we need to make the survival probability decreases as soon as possible.
Thus, we introduced an earliness loss $loss^E$. 
At every time split, the survival probability should be as small as possible. 
For Address $i$ at Time Split $t$, $loss^E$ is defined as:
\begin{equation}
loss^E_{i,t} = -S_{i}(t).
\end{equation}

However, the model is hard to predict the correct labels at the early stage due to data insufficiency. 
The model can be perturbed by the wrong predictions in the early period. 
Thus, all the loss items are weighted by $\sqrt{t}$.
Besides, taking into account that the numbers of positive and reliable negative instances are imbalanced, 
different penalty coefficients are allocated to each class. 
Then, given a set of training samples with $N_p$ malicious addresses and $N_n$ regular addresses,
the overall loss function is defined as:
\begin{equation}
\begin{aligned}
\mathscr{L} = \sum_{t=1}^{t_M}\sqrt{t}(&C^{+}\sum_{i=1}^{N_p}(loss^P_{i,t} + \gamma_{1}{loss^C_{i,t}}+\gamma_{2}{loss^E_{i,t}})+\\
                               &C^{-}\sum_{i=1}^{N_n}(loss^P_{i,t} + \gamma_{1}{loss^C_{i,t}}+\gamma_{2}{loss^E_{i,t}})),
\end{aligned}
\end{equation}
Where $C^{+}$ and $C^{-}$ are inversely proportional to the number of positive and reliable negative instances in our settings. 
$\gamma_{1}$ and $\gamma_{2}$ are coefficients to control the contribution between $loss^P$, $loss^C$ and $loss^E$.

\section{Experiment and Analysis}
\label{sec:experiment}
In this section, we investigate the effectiveness of Path Tracer. First, we introduce the detail of data preparation and evaluation metrics. Next, we perform feature tests to demonstrate the effectiveness of path features. Then, we analyze the effect of the loss function scheme
and 
verify the performance of our path encoder. 
Finally, we provide the contribution ratio of different features at different time steps with visualizations.


\subsection{Data Preparation}
\label{sec:data}
\noindent\textbf{Raw Data and Label Collection}
The transaction data are publicly accessible by running a Bitcoin client. 
As most recent events have not been widely confirmed in the community.
We obtained all the data from the $1$-st block to the $610637$-th block which is the first block of 2020.
We get the related information for a given address based on the APIs exposed by BlockSci~\cite{E1_}.

To get the labels for three different types of malicious addresses (Hack, Ransomware and Darknet), 
we performed a manual search on public forums and datasets, 
such as Bitcointalk forum\footnote{https://bitcointalk.org/}, 
Reddit\footnote{https://www.reddit.com}, 
WalletExplorer\footnote{https://www.walletexplorer.com} and several prior studies \cite{E2_}, \cite{E3_} and \cite{R15_}.
For regular addresses, we collected four types of compliance addresses (Exchange, Mining, Merchant, Gambling) from the same sources (as gambling is legal in some countries).
We call these negative addresses as ``communal negative samples'' which are labeled as negative and shared by all three tasks.

\noindent\textbf{Negative Address Collection}
Since these four types of compliance addresses can not cover all behavior patterns of regular addresses,
we sampled unlabeled addresses at the same time.
Whereas these sampled addresses' labels are not verified, 
%
15\% positive addresses as well as the unlabeled addresses are wrongly labeled as negative addresses.
These 15\% positive addresses are also known as spy instances.
Spy instances are more likely to be predicted as positive than unlabeled addresses.
Thus, we set the probability threshold $\theta$ as the value that can maximize the difference in the growth rates between the cumulative proportion of unlabeled instances and spy instances. 
Addresses with a probability lower than $\theta$ are regarded as reliable negative instances.
The numbers of positive (Posi), unlabeled (Unlab), reliable negative (R-Nega), communal negative (C-Nega) addresses and the Positive/Negative ratio (P/N) for each malicious type (H: Hack, R: Ransomware, D: Darknet) are shown in Table.~\ref{tab:addr_num}.

\begin{table}
    \caption{Dataset Statistics. ``communal negative samples'' are shared by all three tasks.}
    \begin{center}
    \fontsize{10}{11}\selectfont
    \renewcommand{\arraystretch}{1.2}
    \begin{tabular}{|c|c|c|c|c|c|}
    \hline
    Type  & Posi. & Unlab. & R-Nega. &C-Nega. & P/N \cr
    \hline
    \hline
    H   & 341      & 312,443   & 30,673  & 43,092  & 0.46\%\cr
    \hline
    R   & 1,903    &  13,476   & 7,527   & 43,092  & 3.76\%\cr
    \hline
    D   & 7,696    & 543,467   & 46,226  & 43,092  & 8.62\%\cr
    \hline
    \end{tabular}
    \vspace{-0.4cm}
    \label{tab:addr_num}
    \end{center}
    \vspace{-2ex}
\end{table}

\subsection{Settings and Metrics}
\label{sec:metrics}
According to our statistical analysis, $81.3\%$ malicious addresses' behaviors become stable after the first $200$-th hour, and most keep inactive until $1000$-th hour. Therefore, we focus on the first $200$ hours. To evaluate the performance of our model, we get $200$ time steps with $1$ hour interval, and we average the evaluation metrics on all time steps.
The selected metrics are accuracy (Acc.), precision (Prec.), and recall (Rec.).
Besides, we introduce the early-weighted F1 score $F1^{E}$ to evaluate the early detection performance.
We design the consistency-weighted score $F1^{C}$ to evaluate the consistency of the prediction.
\begin{equation}
\begin{aligned}
\mathit{F1^{E}}& = \frac{\sum_{i=1}^{N}{F1_{i}/\sqrt{i}}}{\sum_{i=1}^{N}{1/\sqrt{i}}}, \\
\mathit{F1^{C}}& = \frac{\sum_{i=1}^{N-1}{{\sqrt{i}}\times{F1_{i}}\times{\mathds{1}_{y_c}(y_{i})}}}{\sum_{i=1}^{N-1}{{\sqrt{i}}}},
\end{aligned}
\end{equation}
where $i$ is the time split index, $y_c$ is the prediction set in which $sign(y_i*y_{i+1})>0$. The indicator function $\mathds{1}_{y_c}(y_{i})=1$ when $y_i \in y_c$. $F1_i$ is the $F1$ score of the prediction at the $i$-th time split, denoted as
$y_i$.

\subsection{Feature Combination and Selection}
\label{sec:feat_compare}
\noindent\textbf{Feature Combination}.
To verify the effectiveness of path features, we compare three Decision Tree models, 
namely Address Features (\textbf{AF}) model, Long-Term path features (\textbf{+LT}) model, and Short-Term path features (\textbf{+ST}) model.
\textbf{+X} means add feature \textbf{X} to previous model.
As shown in Table \ref{tab:feat_compare}, \textbf{AF} performed poorly on the three datasets. 
Path features significantly improve performance, especially for the \emph{Recall} and \emph{F1} scores.
We speculate that most of these malicious addresses require victims to transfer money once they are created. 
It makes them quite similar to exchange or financial service addresses in the early stage, as exchange or merchant services will also create new addresses for security. 
As a result, \textbf{AF} model can only find those extremely abnormal addresses, which results in low \emph{Recall} scores. 

\noindent\textbf{Feature Selection}.
Feature selection is crucial for the model's generalization ability, 
to justify the effectiveness of our feature selection scheme, 
we also compare the model's performance under different selection schemes. 
Trimming scheme (\textbf{T}) stands for trimming off features in deletion list, 
Augmentation scheme (\textbf{A}) stands for augmenting features in augmentation list. 
As shown in Table \ref{tab:feat_compare}, feature selection significantly improves the performances of decision trees across all  the datasets. 
The $F1^{C}$ and $F1^{E}$ are enhanced by an average of $13.6\%$ and $12.3\%$, respectively. 
The major improvement comes from the \emph{Recall} score. 
Since malicious activities behave abnormally in various aspects, the most effective features for them are also different. 
Through our automatic feature selection method, models can fully use the powerful path features. 
Also, they can adapt to different malicious activities easily, mitigate the input noise, 
and significantly reduce the workload of manual selection.

\begin{table}[htbp]
\caption{Scores of different features (Address Features (\textbf{AF}), Long-Term path features (\textbf{+LT}), and Short-Term path features (\textbf{+ST})) and selection schemes(Trimming scheme (\textbf{T}), Augmentation scheme (\textbf{A}))}
\begin{center}
\fontsize{10}{11}\selectfont  
\renewcommand{\arraystretch}{1.2}
\begin{tabular}{|c|c|c|c|c|c|c|}
\hline
& $Model$ & $Acc$. & $Prec$. & $Rec$. & $F1^C$ & $F1^E$ \cr
\hline 
\hline
\multirow{5}{*}{{H}} 
                     &AF     &\bf{0.997}	&\bf{0.863}	 &0.164     	 &0.284	         &0.246	\cr
                     &+LT	 &0.996	        &0.755	     &0.154          &0.266	         &0.217	\cr
                     &+ST	 &0.996	        &0.737	     &0.192          &0.303     	 &0.303 \cr
                     \cline{2-7} 
                     &+T	 &\bf{0.997}	&0.833	     &0.191	         &0.310	         &0.301	\cr
                     &+A	 &\bf{0.997}	&0.831	     &\bf{0.207}	 &\bf{0.331}	 &\bf{0.323}\cr
\hline
\multirow{5}{*}{{R}} 
                     &AF     &0.971	        &0.457	     &0.056	         &0.085	         &0.076	\cr
                     &+LT	 &0.972	        &0.620	     &0.125	         &0.205		     &0.202	\cr
                     &+ST	 &0.973      	&\textmd{0.624}   	 &0.220     	 &0.309      	 &0.290\cr
                     \cline{2-7} 
                     &+T	 &0.973	        &\bf{0.627}	 &0.220	         &0.318	         &0.281	\cr
                     &+A	 &\bf{0.974}	&0.604	     &\bf{0.290}	 &\bf{0.387}	 &\bf{0.349}\cr
                     
\hline
\multirow{5}{*}{{D}} 
                     &AF     &0.932         &0.628	     &0.292	         &0.436	         &0.287	\cr
                     &+LT	 &0.928	        &\bf{0.694}	 &0.177	         &0.295	         &0.222	\cr
                     &+ST	 &\bf{0.933}	&0.639	     &0.398	         &0.497	         &0.427\cr
                     \cline{2-7}
                     &+T	 &\bf{0.933}	&0.640   	 &0.390	         &0.473	         &0.425	\cr
                     &+A	 &\bf{0.933}	&0.613	     &\bf{0.455}	 &\bf{0.528}     &\bf{0.470}\cr
                     
\hline
\end{tabular}
\vspace{-2ex}
\label{tab:feat_compare}
\end{center}
\end{table}


\subsection{Prediction Analysis}
\label{sec:predict_ana}
\begin{table*}[t]
    \caption{Performance Comparison with Different Representations and Encoders. Underline stands for best score in the group, Bold stands for best score in all groups.}
    \begin{center}
    \fontsize{8}{11}\selectfont
    \renewcommand{\arraystretch}{1.2}
    \begin{tabular}{|c||c|c|c|c|c|c|c|c|c|c|c|c|c|c|c|}
    \hline
    \multicolumn{16}{|c|}{Datasets} \cr
    \hline
    \multirow{2}{*}{{Model}} &\multicolumn{5}{c|}{Hack} &\multicolumn{5}{c|}{Ransomware} &\multicolumn{5}{c|}{Darknet} \cr
        \cline{2-16}
                             &$Acc$. & $Prec$. & $Rec$. & $F1^C$ & $F1^E$ & $Acc$. & $Prec$. & $Rec$. & $F1^C$ & $F1^E$ & $Acc$. & $Prec$. & $Rec$. & $F1^C$ & $F1^E$ \cr
    \hline
     Feature     &0.984      &0.170      &0.413      &0.308	     &0.194      &0.967      &0.390      &0.511      &0.510      &0.413      &0.818      &0.287      &\ud{0.868} &0.420	     &0.429	\cr
     +Seg.	     &\ud{0.985} &0.176      &\ud{0.442} &0.321      &\ud{0.208} &0.977      &0.633      &0.693      &\ud{0.717} &0.610      &0.848      &0.322      &0.856      &0.434      &0.473\cr
     +Stat.	     &\ud{0.985} &\ud{0.180} &0.441      &\ud{0.327} &\ud{0.208} &\ud{0.980} &\ud{0.674} &\ud{0.716} &0.697      &\ud{0.702} &\ud{0.873} &\ud{0.365} &0.829      &\ud{0.481} &\ud{0.510} \cr
    \hline
     GRU         &\udb{0.997}&0.589      &0.906      &0.698      &0.718      &0.985      &\ud{0.725} &0.964       &0.823      &0.829      &0.906      &0.458      &0.877       &0.569      &0.613	\cr
     M-LSTM	     &\udb{0.997}&0.574      &0.998      &0.725      &0.731      &\ud{0.986} &0.716      &\udb{1.000} &\ud{0.834} &0.835      &\ud{0.907} &\ud{0.465} &0.890       &\ud{0.575} &\ud{0.620}\cr
     SAFE	     &\udb{0.997}&0.610      &\udb{1.000}&\ud{0.758} &\ud{0.757} &0.985      &0.711      &\udb{1.000} &0.831      &0.831      &0.871      &0.363      &0.848       &0.481      &0.512	\cr
     CED         &0.987      &0.339      &0.695      &0.379      &0.474      &\ud{0.986} &0.724      &0.985       &0.827      &\ud{0.837} &0.903      &0.453      &\udb{0.904} &0.565      &0.615	\cr

    \hline
   H-T           &\udb{0.997} &0.615       &\udb{1.000} &0.762       &0.762       &0.986       &0.722       &\udb{1.000} &0.838       &0.838       &0.930       &0.540 &\ud{0.872}      &0.642       &0.674	\cr
   +Surv.	     &\udb{0.997} &0.627       &\udb{1.000} &0.771       &0.770       &0.987       &0.732       &\udb{1.000} &0.844       &0.846       &0.938       &0.585      &0.811      &0.650       &0.688	\cr
   +Loss$_{C,E}$ &\udb{0.997} &\udb{0.634} &\udb{1.000} &\udb{0.776} &\udb{0.776} &\udb{0.988} &\udb{0.747} &\udb{1.000} &\udb{0.855} &\udb{0.855} &\udb{0.941} &\udb{0.594}      &0.833 &\udb{0.668} &\udb{0.701}	\cr
    \hline
    \end{tabular}
    \vspace{0cm}
    \vspace{-0.4cm}
    \label{tab:model_compare}
    \end{center}
\end{table*}

To verify the effectiveness of three representation vectors, we design the following experiments. As shown in the the first group of Table \ref{tab:model_compare},
Model $Feature$ only uses the feature representation. Model $+Seg$ uses the concatenation of feature and segment vectors. Similarly, the $+Stat$ model uses the concatenation of all three kinds of  representations.
We can see, 
on each dataset, $F1^E$ and $F1^C$ have an average improvement of $15.54\%$ and $25.8\%$, proving that our proposed segment and status vectors can significantly improve the model's prediction performance. %
The improvement on the Ransomware dataset is the largest, since ransomware addresses have many obvious stages such as ``address establishing'' or ``waiting for ransom'' 
However, for Hack activities, they are often completed in an instant, most remaining segments have no obvious signals, making the improvement limited.

To verify the validity of our prediction model, we compare 4 SOTA models.
Since all these models are applied in ``Rumor Detection'' task, the characteristics of the input data are different, we modify inapplicable modules for adoption.
As shown in Table \ref{tab:model_compare}, 
(1) \textbf{GRU}~\cite{E8_} is a typical neural network for sequence modeling. We concatenate feature, segment, and status representations at each time split as the input vector to predict the labels with the \textbf{GRU} unit.
(2) \textbf{M-LSTM}~\cite{E6_} adopts LSTM to encode every kind of feature. Here, feature, segment, and status have three independent LSTM models.
(3) \textbf{SAFE}~\cite{E5_} adopts survival probability as the prediction. The positive samples should die out fast, while the negative samples should stay alive.
(4) \textbf{CED}~\cite{E7_} uses RNN for prediction with advanced objective function $O_{diff}$ and $O_{time}$.

In addition, we perform a set of ablation studies to verify modules' effectiveness. 
$H$-$T$ means that we only use the hierarchical transformer encoder. 
$+Surv$ implies using a survival prediction module on top of $H$-$T$.
$+Loss_{C,E}$ means that we consider $Loss_C$ and $Loss_E$ during the training. 

As shown in Table.~\ref{tab:model_compare}, the \textbf{GRU} and \textbf{M-LSTM} models can encode the time series pattern for predictions. 
\textbf{GRU} and \textbf{M-LSTM} have an average improvement of more than $20\%$ in $F1^E$ and $F1^C$ on the three datasets.
This improvement verifies the effectiveness of temporal patterns.

In addition, a pre-stopping mechanism can eliminate the interference of useless noise. For this, we compare \textbf{SAFE} and \textbf{CED} models.
With the introduction of pre-stopping mechanism, they achieve better performances in some datasets.
However, the \textbf{SAFE} model has a monotonically increasing positive hazard rate, the previous lousy decision can never be modified, resulting in lower precision scores.
Besides, \textbf{CED} models unable to use previous predictions directly, thus cannot guarantee the prediction consistency.
Also, they do not have a hierarchical structure design between features, segments, and status. 
Different malicious activities have different interactions between feature, segment, and status. 
Ignoring the inherent hierarchical relationship will weaken the early stopping module's generalization ability. 
Thus, their performances are not stable among these three datasets.

As shown in Table.~\ref{tab:model_compare}, compared to \textbf{SAFE} and \textbf{CED}, the $H$-$T$ model can achieve an average improvement of $5.00\%$ and $4.92\%$ on $F1^C$ and $F1^E$,
proving that the hierarchical relationship of the three representations can improve the prediction performance.
We can improve the prediction by introducing survival probability based on the $H$-$T$ model. Compared with the $H$-$T$ model, 
$F1^C$ and $F1^E$ are increased by an average of $1.04\%$ and $1.36\%$. 
This increment suggests that the survival module accelerates the model's prediction speed and improves the accuracy of the model prediction, thereby improving $F1^C$. 
Finally, the model was again enhanced by $1.57\%$ and $1.24\%$ through Early Loss and Predict Consistent Loss based on the previous step. 
The final increases of $F1^C$ and $F1^E$ are $7.83\%$ and $6.21\%$ compared with the best scores among \textbf{SAFE} and \textbf{CED}.

\subsection{Feature Contribution}
\label{sec:feat_contri}
\begin{figure}
	\centering
	\vspace{-0ex}
	\includegraphics[width=.9\columnwidth, angle=0]{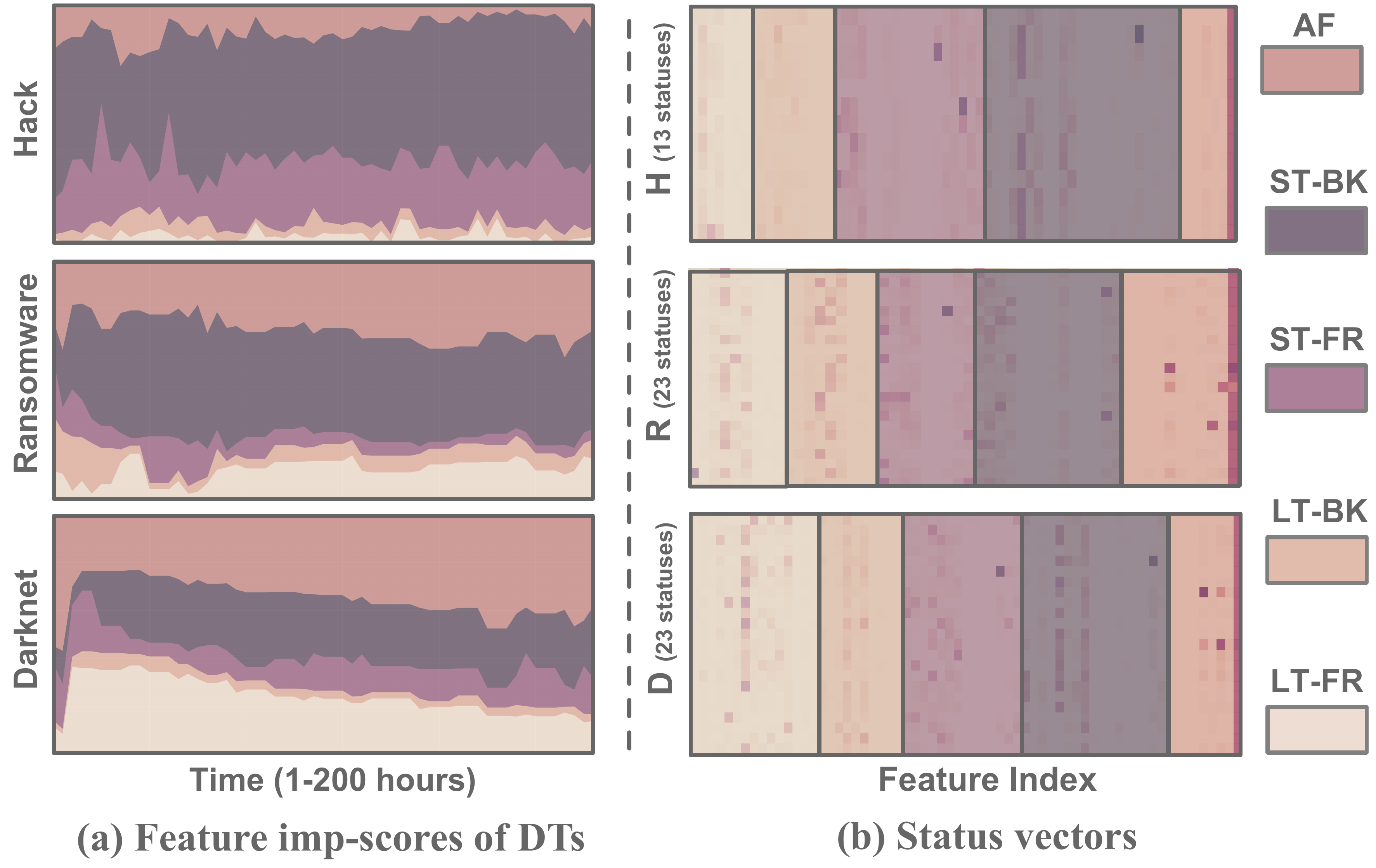}
	\vspace{-1ex}
	\caption{Feature importance score evolution and feature constitution for status vectors.}
	\label{fig:hack_status}
	\vspace{-5ex}
\end{figure}
To verify that our model selects features differently for different activities, we plot the importance score over time.
First, as shown in Fig.~\ref{fig:hack_status}, we see that the Path feature has high importance scores on all the three datasets, which again proves the usefulness of our path feature.
In addition, we found that for Hack and Ransomware, the ST-Path feature has a higher contribution than the LT-Path feature.

For hack, it is generally a criminal who invades the system of an exchange or a large company, steals the private key of its wallet, and then quickly transfers the funds to a secure account through a series of methods such as peeling-chain. Since the fund's source is the exchange with tensive connections with outside, the ST-BK path provides rich temporal and topology information, so ST-BK accounts for the most significant proportion. Secondly, money laundering activities will gradually disperse dirty money so that the transfer-out structural pattern can be well reflected in the ST-FR path.

For Ransomware, the source of funds is similar to a Hack. The victims are usually exchanges, or pay ransoms through exchanges, so the weight of ST-BK's path will be higher. Because there will be some time lag for the victim to pay the ransom, the information can usually be reflected by the Address-Features (AF). So the weight of the AF is also relatively high. In addition, since Ransomware usually has multiple victims, there will be various addresses accepting the ransom, and each address will aggregate all the ransoms into one total address. Therefore, the weight of LT-FR will be higher.

For Darknet, its nature is similar to a trading platform, but due to the concealment, users are required to pay tokens in a specific way, leading to a slightly higher proportion of ST-BK. In addition, because it is a platform, the need for money laundering is not too frequent, so the pattern of LT-FR can help distinguish Darknet addresses from other regular addresses.

For importance score evolution, the trend of Hack and Ransomware will not change much because these two behaviors last for a short time. After completing the task, there will be no further visible activity in these early 200 hours. On the other hand, Darknet will continue to operate and get exposed gradually over time, so the weight of AF will slowly increase.

\begin{figure}
	\centering
	\vspace{-0ex}
	\includegraphics[width=1.\columnwidth, angle=0]{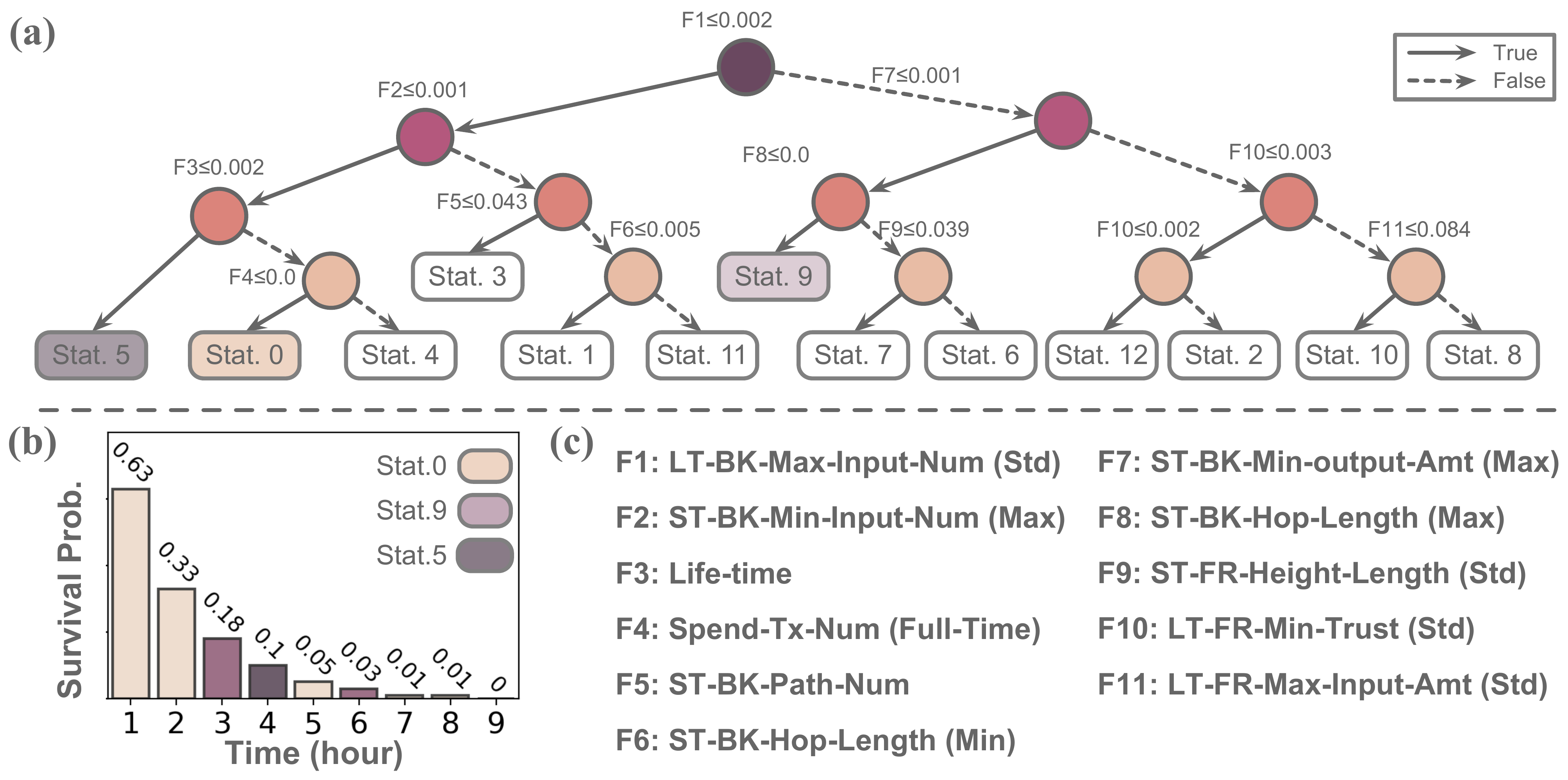}
	\vspace{-4ex}
	\caption{(a) Status decision tree for Hack dataset. 
	         (b) Survival Probability of example address. Color are consistent with leaf nodes in status decision tree.
	         (c) Semantic meaning of features in decision tree}
	\label{fig:hack_status}
	\vspace{-2ex}
\end{figure}

\subsection{Status and Intention}
\label{sec:stat_and_inten}
\begin{figure}
	\centering
	\vspace{-0ex}
	\includegraphics[width=1.\columnwidth, angle=0]{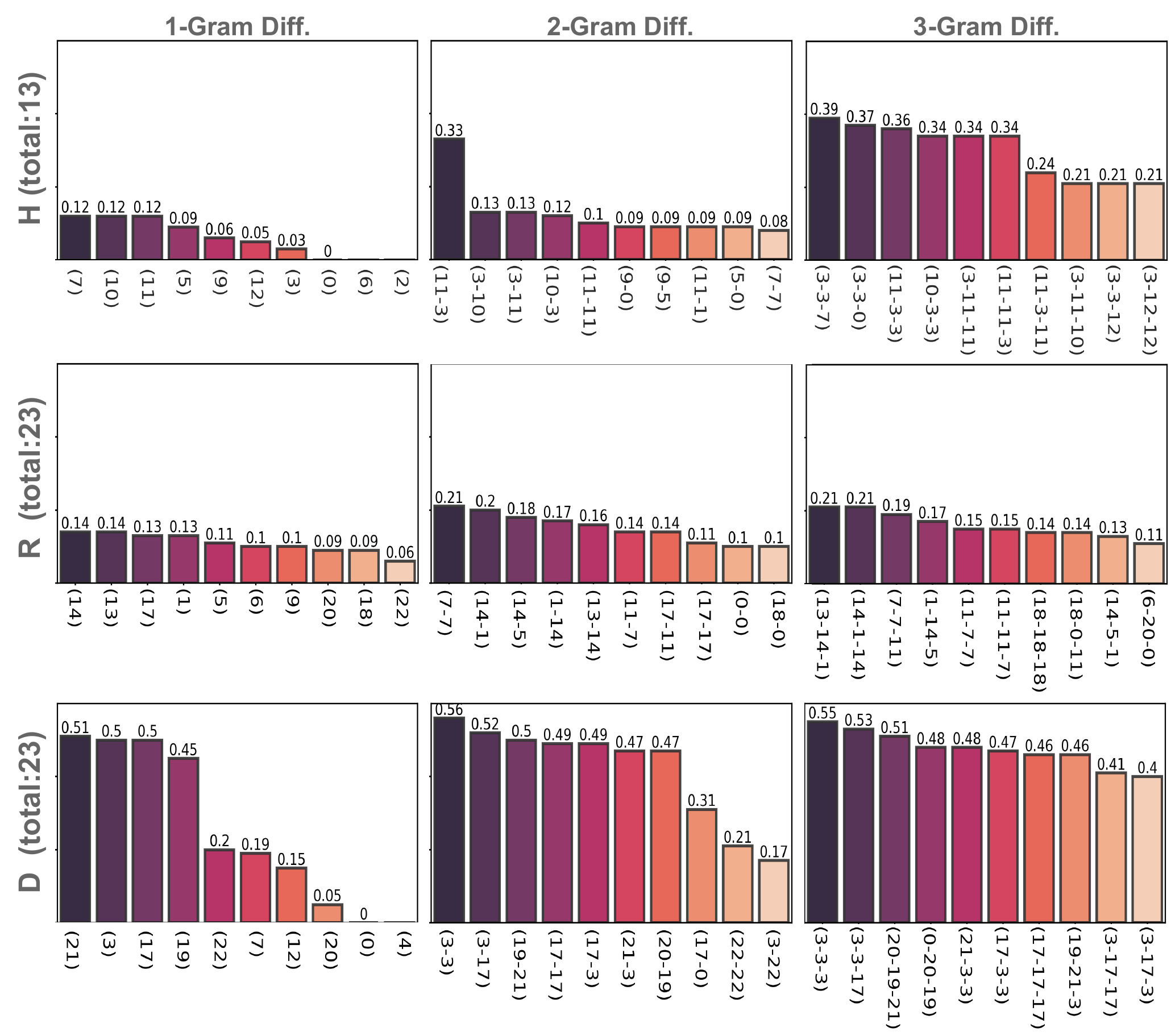}
	\vspace{-4ex}
	\caption{Top-10 differentiated status n-gram. Labels on x-axis stand for the names of status n-gram. Value above each bar is the proportion difference between malicious and regular addresses.}
	\label{fig:stat_prop_diff}
	\vspace{-4ex}
\end{figure}

\noindent\textbf{N-Gram Analysis}.
%
We pay more attention to the statuses that appear more frequently on malicious addresses. We select the top ten with the largest differences for analysis. As shown in Fig.~\ref{fig:stat_prop_diff}, for single status (1-Gram), only a few statuses in malicious addresses have a higher proportion than regular addresses, and the discrimination is relatively small. But when combined into 2-grams or 3-grams, the difference becomes apparent. So we argue that the status reflects a general state. An address can stay at any status, but the status sequence reflects different addresses' intentions which is crucial for early detection of malicious addresses.

\noindent\textbf{Case Study}.
We use the infamous hack incident of Binance crypto exchange on May 7, 2019 \footnote{https://www.cnbc.com/2019/05/08/binance-bitcoin-hack-over-40-million-of-cryptocurrency-stolen.html} in which hackers stole over 7000 bitcoins worth of 40 million, as a case study to illustrate our model's capability to better interpret prediction result and offer useful insights into the malicious behavior. 

Our method successfully detects the hacker address involved by the end of the first 9 hours since its creation, which is \emph{12 hours before the stolen bitcoins were transferred away}. We quote from a Binance statement that "It was unfortunate that we were not able to block this withdrawal before it was executed.", an inevitable tragedy with only retrospective analysis but totally preventable with our early detection.   

To interpret our prediction result, 
 we first zoom into one of the hacker addresses bc1***3wm\footnote{bc1q8m9h3atn4cqeqhu3ekswdqxchp3g7d4v3qv3wm}. 
As shown in Fig.~\ref{fig:hack_status} (b), the status-sequence of the first 9 hours is [0-0-9-5-0-9-0-0-0], which is generated as a witness intention motif when our model correctly predicts the malicious address as
the survival probability decreases below our threshold by the $9$-th hour. 

Our prediction result can be easily interpreted by examining the semantics behind the statuses that form the intention motif. Each status can be interpreted by conducting a tree traversal on the status decision tree from the root to the leaf corresponding to the status as shown in Fig.~\ref{fig:hack_status} (a).

\textbf{Status 0} essentially indicates (I) The asset associated with the address has been obtained from a single source. (A small F1 value means the address has at most one LT-BK path and a small F2 value means most ST-BK paths are single chains.), and (II) There no spend transaction (A small F4 means no transaction).
\textbf{Status 9} reveals a unique pattern by which the address receives the asset -- the asset has been obtained from a single source yet through a sequence of asset transitions each of which "peels" a tiny amount off the whole asset before passing onto the receiver, a highly suspicious pattern similar to money laundering where asset are transitioned through a long chain of accounts. 
\textbf{Status 5} is similar to status \textbf{Status 0}, only measured at a later time-stamp, to confirm that there has been no spend transaction after the initial asset received from a single source at the early beginning.

The combination of these statuses as the witness intention motif provides sufficient evidence for our model to predict the malicious address by the end of the $9$-th hour. What actually happened in reality is that, this malicious address received a transfer of $567.997$ BTCs at creation through $71$ input transitions, with no output transactions. By the end of $13$-th hour, it received another transfer about $0.00008642$ BTC, followed by a bulk transfer out of all its BTC asset at $21$st hour. Indeed, by the end of $13$-th hour our model would be fully confident of the prediction as the survival probability decreases to literally 0 by $13$-th hour.

\begin{figure}
	\centering
	\vspace{-0ex}
	\includegraphics[width=1.\columnwidth, angle=0]{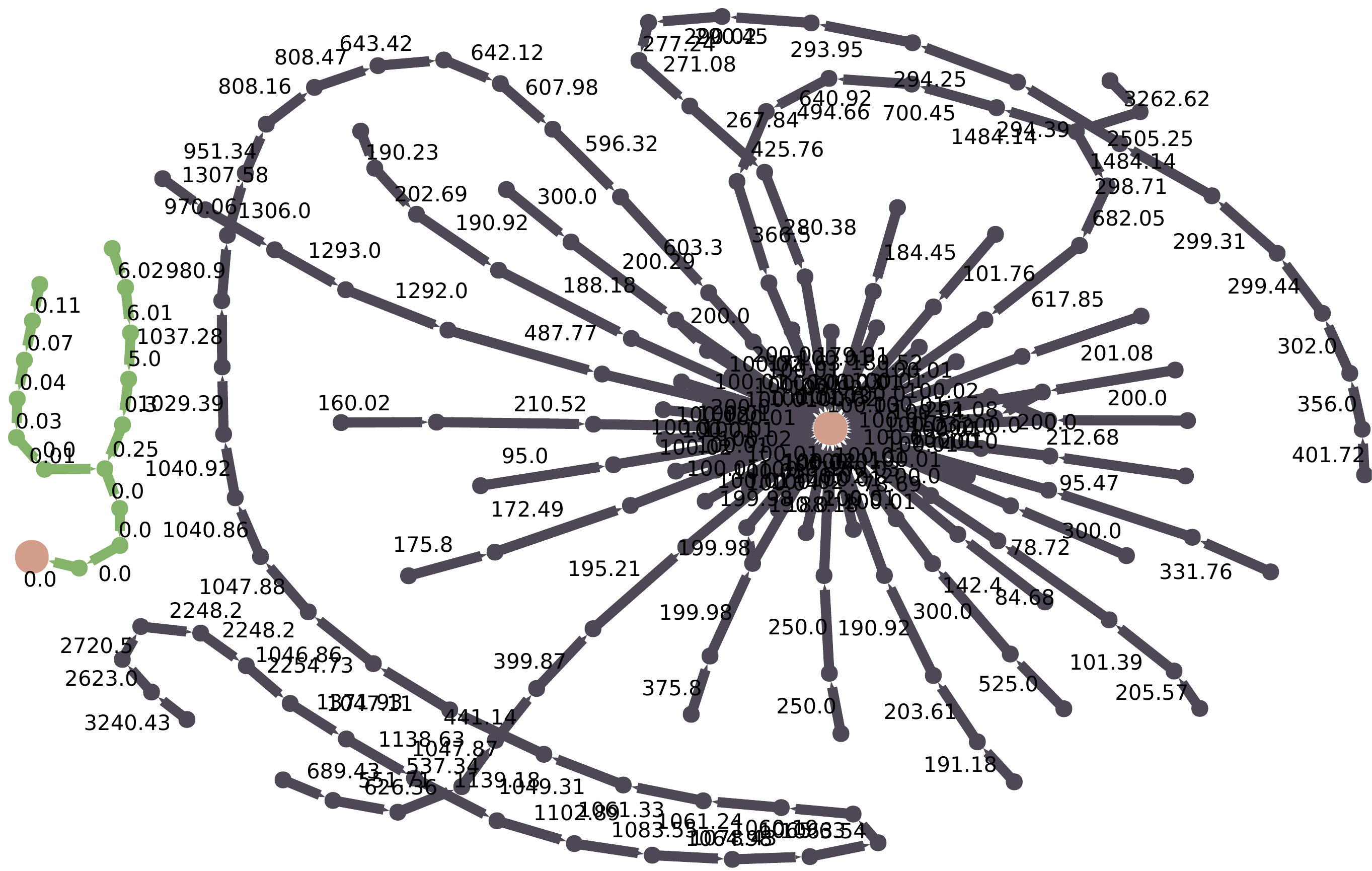}
	\vspace{-2ex}
	\caption{Short-Term Backward (ST-BK) paths for the instance address in the first $24$ hours (gray and green edges stand for ST-BK paths of the address's first and second incoming transactions.). The numbers stand for the amount of the transaction edges. The pink nodes are instance related transaction.}
	\label{fig:case_ST_BK}
	\vspace{-4ex}
\end{figure}

Besides this particular hacker address discussed, this witness intention motif of [0-0-9-5-0-9-0-0-0] can also be observed in the related transaction\footnote{e8b406091959700dbffcff30a60b190133721e5c39e89bb5fe23c5a554ab05ea}, in which hackers manipulated Binance's address and divided it into $71$ inputs, each containing 100 BTCs. The pattern can be clearly seen in Fig.~\ref{fig:case_ST_BK}, those black edges are the ST-BK paths related to the first input transfer.
It is evident from this case study that
our method could achieve early detection of malicious addresses to thwart their illicit campaign before it is too late. 
Our prediction results are also offered with good interpretation by examining the semantics behind the witness intention motif generated together with the prediction. 

Moreover, useful insights can be acquired from our prediction result. For example, the extremely tiny amount of $0.00008642$ BTC received by the malicious address by the end of $13$-th hour is highly likely the corroborating evidence that it is a trial transfer to test whether the transfer operation is successful, as a specific signal transaction to automatically coordinate and synchronize multiple addresses' hacking operations. This was also validated by Binance statment in which they pointed out that “The hackers had the patience to wait, and execute well-orchestrated actions through multiple seemingly independent accounts at the most opportune time.”

As another example, although our model has already given the prediction in the $9$-th hour, combined with the subsequent status of our proposed ST-BK path, we can even identify potentially a group of hackers for this hack incident.
As shown in the Fig.~\ref{fig:case_ST_BK}, the green edge is a signal transaction after $13$ hours, and the amount on it is tiny but introduces one LT-BK path and two ST-BK Paths. These two ST-BK Paths merged into a single track before importing to the hack address.


We back-tracked the source of the ST-BK path of the signal transaction. We found that the $21$ hack addresses that participated in this hack incident were linked through the signal transactions. Even more surprisingly, they have the same source coming from the address\footnote{1GrdXZpyBfiNSVereX5t5UQRHfeh192Cc6}. Our belief that multiple addresses launched this hacking and synchronized among themselves through signal transactions again echoes the collaborative schemes claimed in the Binance statement. 
%
%
%

\section{Conclusion}
\label{sec:conclusion}
This paper presents Intention Monitor, a novel framework for the early detection of malicious addresses on BTC. 
After proposing two kinds of asset transfer paths, 
we select, augment, and split the feature sequence for different malicious activities with a decision tree-based strategy. 
In particular, we propose segment and status vectors to describe the temporal behaviors and global semantic status.
We build a hierarchical transformer encoder to capture the semantic-level relation among them to take full advantage of three representations. 
In particular, a survival prediction module and corresponding loss items facilitate a faster and consistent prediction.
We quantitatively and qualitatively evaluated the model on three malicious address datasets.
Extensive ablation studies were conducted to determine the mechanisms behind model's effectiveness. 
The experimental results show that the proposed method outperformed the state-of-the-art baseline approaches on all three datasets.
Furthermore, a detailed case study on Binance Hack justifies that our model can not only explain the suspicious transaction patterns but can also find hidden abnormal signals.

\bibliography{reference}
\bibliographystyle{IEEEtran}

\end{document}